%% file: MainFile.tex
\documentclass[journal,comsoc]{IEEEtran}
\usepackage{epsfig}
\usepackage{graphicx}
\usepackage{amsmath}
\usepackage{amssymb}
\usepackage{multirow}
\usepackage{tabulary}
\usepackage{tabularx}
\usepackage[tight,normalsize,sf,SF]{subfigure}
\usepackage{color}
\usepackage{soul}
\usepackage{enumitem}
\usepackage{hhline}
\usepackage{algorithm}
\usepackage{algpseudocode}
\usepackage{array}
\usepackage{flushend}
\usepackage{multirow}
\usepackage{tabulary}
\usepackage{tabularx}
\usepackage{bm}
\usepackage[tight,normalsize,sf,SF]{subfigure}
\usepackage{color}
\usepackage{csquotes}
\DeclareMathSymbol{\Lambda}{\mathalpha}{operators}{3}
\DeclareMathSymbol{\Pi}{\mathalpha}{operators}{5}
\newcolumntype{P}[1]{>{\centering\arraybackslash}p{#1}}
\graphicspath{{./figs/}}
\DeclareGraphicsExtensions{.eps,.ps,.pdf,.png,.tiff}
\renewcommand{\baselinestretch}{0.98}

\usepackage[T1]{fontenc}
\begin{document}

\title{Multi-View Surveillance Video Summarization via Joint Embedding and Sparse Optimization}

\author{Rameswar~Panda
        and~Amit~K.~Roy-Chowdhury,~\IEEEmembership{Senior~Member,~IEEE}
\thanks{$\bullet$ Rameswar Panda and Amit~K.~Roy-Chowdhury are with the Department of Electrical and Computer Engineering, University of California, Riverside, CA 925024, USA. 
E-mails: (rpand002@ucr.edu, amitrc@ece.ucr.edu)}
}

\markboth{IEEE Transactions on Multimedia,~Vol.~xx, No.~xx, December~20xx}%
{Shell \MakeLowercase{\textit{et al.}}: Bare Demo of IEEEtran.cls for IEEE Communications Society Journals}

\maketitle

\begin{abstract}
Most traditional video summarization methods are designed
to generate effective summaries for single-view
videos, and thus they cannot fully exploit the complicated
intra and inter-view correlations in summarizing multi-view videos in a camera network.
In this paper, with the aim of summarizing multi-view videos, we introduce a novel unsupervised framework via joint embedding and sparse representative selection. The objective function is two-fold. The first is to capture the multi-view correlations via an embedding, which helps in extracting a diverse set of representatives. 
The second is to use a $\ell_{2,1}$-norm to model the sparsity while selecting representative shots for the summary. We propose to jointly optimize both of
the objectives, such that embedding can not only characterize the correlations, but also indicate the requirements of
sparse representative selection. We present an efficient alternating algorithm based on half-quadratic minimization to solve the proposed non-smooth and
non-convex objective with convergence analysis. 
A key advantage of the proposed approach with respect to the state-of-the-art is that it can summarize multi-view videos without assuming any prior correspondences/alignment between them, e.g., uncalibrated camera networks.   
Rigorous experiments on several multi-view
datasets demonstrate that our approach clearly
outperforms the state-of-the-art methods. 
       
\end{abstract}


\IEEEpeerreviewmaketitle

\section{Introduction}
\label{sec:Introduction}
\input{Introduction}

\vspace{-3mm}
\section{Related Work}
\label{sec:Related Work}
\input{RelatedWork}
\vspace{-4mm}
\section{Proposed Methodology}
\label{sec:Multi-Video Summarization}
\input{Methodology}

\vspace{-2mm}
\section{Experiments}
\label{sec:Experiments}
\input{Experiments}

\vspace{-1mm}
\section{Conclusions and Future Works}
\label{sec:Conclusions}
\input{Conclusions}

\section*{Acknowledgment}

This work was partially supported by NSF grant 1544969. We gratefully acknowledge the
support of NVIDIA with the donation of the
Tesla K40 GPU used for this research.  

{
	\bibliographystyle{ieee}
	\bibliography{egbib}
}

\end{document}

%% file: Introduction.tex
\IEEEPARstart{N}{etwork} of surveillance cameras are everywhere nowadays. The volume of data collected by such network of vision sensors deployed in many settings ranging from security needs to environmental monitoring clearly meets the requirements of big data~\cite{huang2014surveillance,roy2012camera}. The difficulties in analyzing and processing such big video data is apparent whenever there is an incident that requires foraging through vast video archives to identify events of interest.
As a result, \textit{video summarization}, that automatically extract a brief yet informative summary of these videos has attracted intense attention in the recent years.

\begin{figure}[!t]
	\centering
	\begin{tabular}{c}
		\includegraphics[scale=0.40]{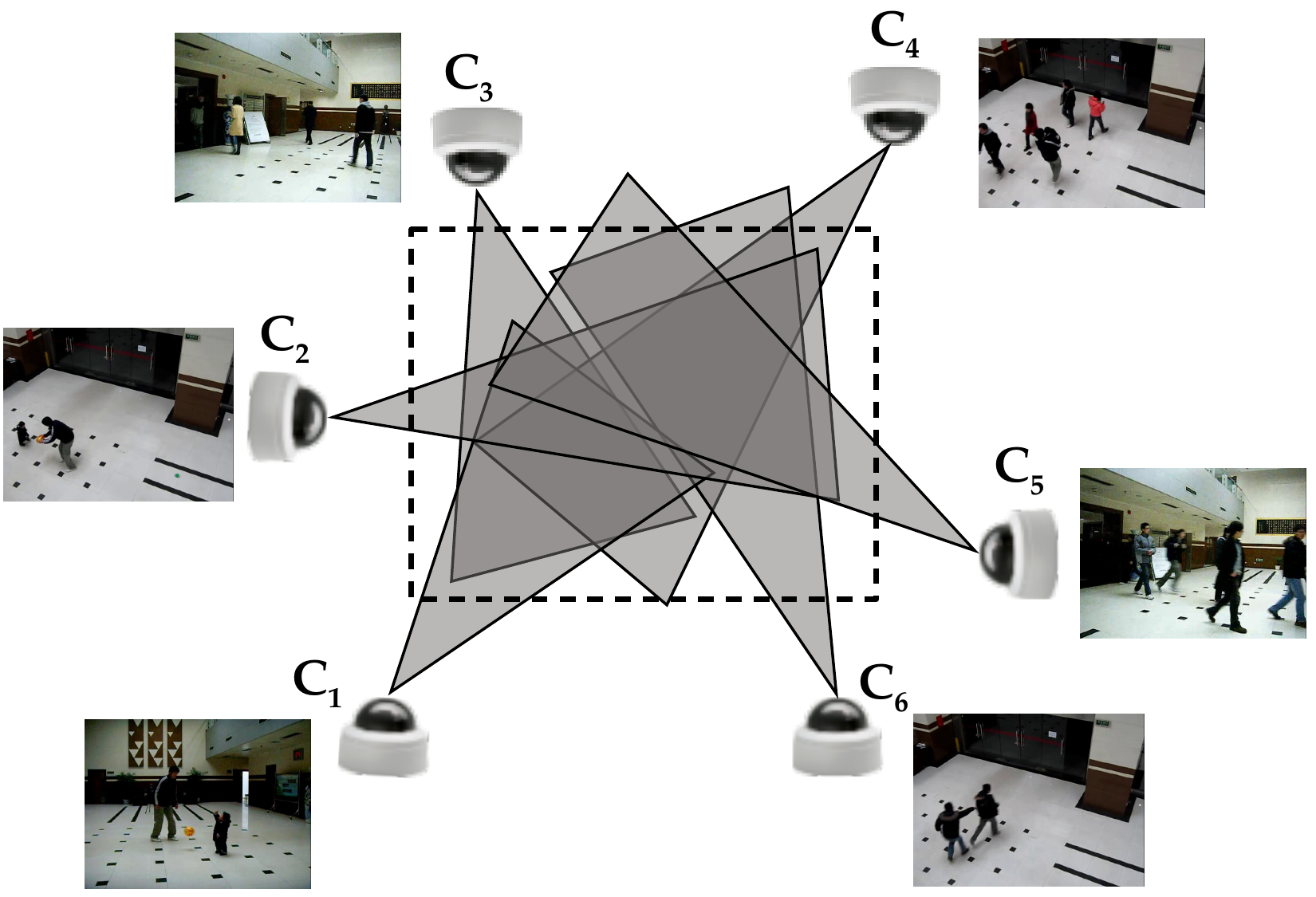}
	\end{tabular}
	\caption
		{An illustration of a multi-view camera network where six cameras $C_1$, $C_2$, \dots, $C_6$ are observing an area (black rectangle) from different viewpoints. 
		Since the views are roughly overlapping, information correlations across multiple views along with correlations in each view should be taken into account for generating a concise multi-view summary.    
	}
	\label{fig:ProblemFig} \vspace{-3mm}
\end{figure}

Although video summarization has been extensively studied
during the past few years, many previous methods mainly
focused on developing a variety of ways to summarize \textit{single-view} videos in form of a key-frame sequence or a video skim~\cite{Category2014,Ehsan2012,Eric2014,VSUMM2011,Joint2014,Khosla2013,JVCI2013}. However, another important problem and rarely addressed in this context is to find an informative summary from \textit{multi-view} videos~\cite{MultiviewTMM2010,MultiviewICIP2011,OnlineMultiview2015,SanjaySir2015,panda2016embedded}. \textit{Multi-view video summarization refers to the problem of summarization that seeks to take a set of input videos captured from different cameras focusing on roughly the same fields-of-view (fov) from different viewpoints and produce a  video synopsis or key-frame sequence that presents the most important portions of the inputs within a short duration (See Fig.~\ref{fig:ProblemFig}).} In this paper, given a set of videos and its shots, we focus on developing an unsupervised approach for selecting a subset of shots that constitute the multi-view summary. Such a summary can be very beneficial in many surveillance systems equipped in offices, banks, factories, and crossroads of cities, for obtaining significant information in short time.

Multi-view video summarization is different from single-video summarization in two important ways. First, although the amount of multi-view data is immensely challenging, there is a certain structure underlying it. Specifically, there is large amount of correlations in the data due to the locations and fields of view of the cameras. So, content correlations as well as discrepancies among different videos need to be properly modeled for obtaining an informative summary. 
Second, these videos are captured with different view angles, and depth of fields, for the same scenery, resulting in a number of unaligned videos. Hence, difference in illumination, pose, view angle and synchronization issues pose a great challenge in summarizing these videos. So, methods that attempt to extract summary from single-view videos usually do not produce an optimal set of representatives while summarizing multi-view videos.

To address the challenges encountered in a camera network, we propose a novel multi-view video summarization method, which has the following advantages.
\begin{itemize}[leftmargin=*] 
	\item First, to better characterize the multi-view structure, we project the data points into a latent embedding which is able to preserve both intra and inter-view correlations without assuming any prior correspondences/alignment between the multi-view videos, e.g., uncalibrated camera networks.
	Our underlying idea hinges upon the basic concept of subspace learning~\cite{chennubhotla2001sparse,nguyen2012sparse}, which typically aims to obtain a latent subspace shared by multiple views by assuming that these views are generated from this subspace.
	
	\item Second, we propose a sparse representative selection method over the learned embedding to summarize the multi-view videos. Specifically, we formulate the task of finding summaries as a sparse coding problem where the dictionary is constrained to have a fixed basis (dictionary to be the matrix of same data points) and the nonzero rows of sparse coefficient matrix represent the multi-view summaries.
	
	\item Finally, to better leverage the multi-view embedding and selection mechanism, we learn the embedding and optimal representatives jointly. Specifically, instead of simply using the embedding to characterize multi-view correlations and then selection method, we propose to adaptively change the embedding with respect to the representative selection mechanism and unify these two objectives in forming a joint optimization problem. With joint embedding and sparse representative selection, our final objective function is both non-smooth and non-convex. We present an efficient optimization algorithm based on half-quadratic function theory to solve the final objective function. 
\end{itemize}

%% file: RelatedWork.tex
There is a rich body of literature in multimedia and computer vision on summarizing videos in form of a key frame sequence or a video skim (see~\cite{money2008video,Truong2007} for reviews).  

\underline{\it Single-view Video Summarization.} Much progress has been made in developing a variety of ways to summarize a single-view video in an unsupervised manner or developing supervised algorithms. Various strategies have been studied, including clustering~\cite{VISON2012,VSUMM2011,Top2014,Salehin2016}, attention modeling~\cite{Att2005}, saliency based linear regression model~\cite{Graumann2012}, super frame segmentation~\cite{LucVanGool2014}, kernel temporal segmentation~\cite{Category2014}, crowd-sourcing~\cite{Khosla2013}, energy minimization~\cite{Peleg2007,Onlinecontent2012}, storyline graphs~\cite{Joint2014}, submodular maximization~\cite{gygli2015video}, determinantal point process~\cite{gong2014diverse,zhang2016summary}, archetypal analysis~\cite{song2015tvsum}, long short-term memory~\cite{zhang2016video} and maximal biclique finding~\cite{chu2015video}.

Recently, there has been a growing interest in using sparse coding (SC) to
solve the problem of video summarization~\cite{Ehsan2012,Eric2014,Scalable2012,meng2016keyframes,dornaika2015decremental,mei2015video} since the sparsity and reconstruction error term naturally fits into the problem of summarization. 
In contrast to these prior works that can only summarize a single video, we develop a novel multi-view summarization method that jointly summarizes a set of videos to find a single summary for describing the collection altogether.

\underline{\it Multi-view Video Summarization.} Generating a summary from multi-view videos is a more challenging problem due to the inevitable thematic diversity and content overlaps within multi-view videos than a single video. 
To address the challenges encountered in multi-view settings, there have been some specifically designed approaches that use random walk over spatio-temporal graphs~\cite{MultiviewTMM2010} and rough sets~\cite{MultiviewICIP2011} to summarize multi-view videos. A recent work in~\cite{SanjaySir2015} uses bipartite matching constrained optimum path forest clustering to solve the problem of multi-view video summarization. An online method can also be found in~\cite{OnlineMultiview2015}. However, this method relies on inter-camera frame correspondence, which can be a very difficult problem in uncontrolled settings.  The work in ~\cite{ManjunathAccv2011} and ~\cite{ManjunathTON2014} also addresses a similar problem of summarization in non-overlapping camera networks. 
Learning from multiple information sources such as video tags~\cite{wang2012event}, topic-related web videos~\cite{panda2017collaborative,panda2016sparse} and non-visual data~\cite{zhu2016learning,wang2013amazon} is also a recent trend in multiple web video summarization.

This paper has significant differences with our previous work in~\cite{panda2016video}. First, in~\cite{panda2016video}, we proposed a static multi-view summarization method that extracts a set of key frames to present most important portions of the input videos in form of story-boards. While key frames are a helpful way of indexing videos, they are limited in that all motion information is lost. This limits their use in many surveillance applications where video skimming i.e., replacing all the videos by a shorter compilation
of its fragments/shots, seems better suited for obtaining
significant information in short time. In this work, we focus on dynamic shot-based video summarization, which not only reduces computational cost but also provides more flexible way of representing videos by considering temporal aspects of activities typically shown in videos. Towards this, we propose a video representation scheme based on spatio-temporal C3D features which have recently shown promising results in several video recognition tasks~\cite{tran2014c3d,panda2017collaborative}. Second, in~\cite{panda2016video}, we adopt a two step approach i.e., both embedding and representative selection are performed independently while summarizing multi-view videos. By contrast, in this work, we jointly optimize both of the objectives, such that the embedding can not only characterizes the multi-view structural correlations, but also indicates the requirements of sparse representative selection. Experiments show that joint optimization indeed improves the summarization performance by generating more informative multi-view summaries. Third, we conduct rigorous experiments on three additional multi-view datasets including one large scale dataset captured with 19 surveillance cameras in an indoor setting~\cite{OnlineMultiview2015}. We also perform a subjective user study to validate the effectiveness of our approach in generating high quality summaries for a more efficient and engaging viewing experience. New experimentation with spatio-temporal video representation, and the joint optimization well demonstrate the performance improvements in the current framework for summarizing multi-view videos.

%% file: Methodology.tex
In this section, we start by giving main notations and definition of the multi-view summarization problem and then present our detailed approach to summarize multi-view videos.

\vspace{1mm}

\underline{\it Notation.} We use uppercase letters to denote matrices and lowercase letters to denote vectors. For matrix $A$, its $i$-th row and $j$-th column are denoted by $a^i$ and $a_j$ respectively. $||A||_F$ is Frobenius norm of $A$ and $tr(A)$ denote the trace of $A$. The $\ell_p$-norm of the vector $a \in \mathbb{R}^n$ is defined as $||a||_p = (\sum_{i=1}^{n}|a_i|^p)^{1/p}$ and $\ell_0$-norm is defined as $||a||_0 = \sum_{i=1}^{n}|a_i|^0$. The $\ell_{2,1}$-norm can be generalized to $\ell_{r,p}$-norm which is defined as $||A||_{r,p}= (\sum_{i=1}^{n}||a^i||_{r}^{p})^{1/p}$. 
The operator $diag(.)$ puts a vector on the main diagonal of a matrix.

\vspace{1mm}

\underline{\it Multi-View Video Summarization.} Given a set of videos captured with considerable overlapping fields-of-view across multiple cameras, the goal of multi-view video summarization is to compactly depict the input videos, distilling its most informative events into a short watchable synopsis. Specifically, it is composed of several video shots that represent most important portions of the input video collection within a short duration.     

\vspace{1mm}

Our approach can be roughly described as the set of three main tasks, namely (i) video representation, (ii) joint embedding and representative selection, and (iii) summary generation. In particular, our approach works as follows. First, we segment each video into multiple non-uniform shots using an existing
temporal segmentation algorithm and represent each shot by a feature vector using a mean pooling scheme over the extracted C3D features (Section~\ref{sec:video representation}). Then, we develop
a novel scheme for joint embedding and representative selection by exploiting the multi-view correlations without assuming any prior correspondence between the videos (Sections~\ref{sec:Embedding}, ~\ref{sec:Summarization}, ~\ref{sec:Joint}). Specifically, we formulate the task of finding summaries as an $\ell_{2,1}$ sparse optimization where the nonzero rows of sparse coefficient matrix represent the relative importance of the corresponding shots.
Finally, the approach outputs a video summary composed of the shots with the highest importance score (Section~\ref{sec:sumg}). 
\vspace{-3mm}
\subsection{Video Representation}
\label{sec:video representation}
Video representation is a crucial step in summarization for maintaining visual coherence, which in turn affects the overall quality of a summary. It basically consists of two main steps, namely, (i) temporal segmentation, and, (ii) feature representation. We describe these steps in the following sections. 

\vspace{1mm}

\underline{\it Temporal Segmentation.} 
Our approach starts with segmenting videos using an existing algorithm~\cite{chu2015video}. We segment each video into multiple shots by measuring the amount of changes between two consecutive frames in the RGB and HSV color spaces. A shot boundary is determined at a certain frame when the portion of total change is greater than 75\%~\cite{chu2015video}. We added an additional constraint to the algorithm to ensure that the number of frames within each shot lies in the range of [32,96]. The segmented shots serve as the basic units for feature extraction and subsequent processing to extract a summary.

\vspace{1mm}

\underline{\it Feature Representation.} Recent advancement in deep feature learning has revealed that features extracted from upper or intermediate layers of a CNN are generic features that have good transfer learning capabilities across different domains~\cite{simonyan2014two,karpathy2014large}.
An advantage of using deep learning features is that there exist accurate, large-scale datasets such as Imagenet~\cite{russakovsky2015imagenet} and Sports-1M~\cite{karpathy2014large} from which they can be extracted.
For videos, C3D features~\cite{tran2014c3d} have recently shown better performance compared to the features extracted using each frame separately~\cite{tran2014c3d,yao2015describing}.
We therefore extract C3D features, by taking sets of 16 input frames, applying 3D convolutional filters, and extracting the responses at layer FC6 as suggested in~\cite{tran2014c3d}. 
This is followed by a temporal mean pooling scheme to maintain the local ordering structure within a shot. 
Then the pooling result serves as the final feature vector of a shot (4096 dimensional) to be used in the sparse optimization. We will discuss the performance benefits of employing C3D features in our experiments.
\vspace{-2mm} 
\subsection{Multi-view Video Embedding}
\label{sec:Embedding}
Consider a set of $K$ different videos captured from different cameras, where ${X}^{(k)} = \{x_i^{(k)} \in \mathbb{R}^D ,i = 1,\cdots,N_k\}, k = 1,\cdots,K$. 
Each $x_i$ represents the feature descriptor of a video shot in $D$-dimensional feature space. We represent each shot by extracting the shot-level C3D features as described above. 
As the videos are captured non-synchronously, the number of shots in each video might be different and hence there is no optimal one-to-one correspondence that can be assumed. 
We use $N_k$ to denote the number of shots in $k$-th video and $N$ to denote the total number of shots in all videos. 

Given the multi-view videos, our goal is to find an embedding for all the shots into a joint latent space while satisfying some constraints. 
Specifically, we are seeking a set of embedded coordinates ${Y}^{(k)} = \{y_i^{(k)} \in \mathbb{R}^d ,i = 1,\cdots,N_k\}, k = 1,\cdots,K$, where, $d$ $(<<D)$ is the dimensionality of the embedding space, with the following two constraints: {(1)} \textit{Intra-view correlations.} The content correlations between shots of a video should be preserved in the embedding space. {(2)} \textit{Inter-view correlations.} The shots from different videos with high feature similarity should be close to each other in the resulting embedding space as long as they do not violate the intra-view correlations present in an individual view.

\underline{\it Modeling Multi-view Correlations.} To achieve an embedding that preserves the above two constraints, we need to consider feature similarities between two shots in an individual video as well as across two different videos.

Inspired by the recent success of sparse representation coefficient based methods to compute data similarities in subspace clustering~\cite{elhamifar2013sparse1}, we adopt such coefficients in modeling multi-view correlations.
Our proposed approach has two nice properties: (1) the similarities computed via sparse coefficients are robust against noise and outliers since the value not only depends on the two shots, but also depends on other shots that belong to the same subspace, and (2) it simultaneously carries out the adjacency construction and similarity calculation within one step unlike kernel based methods that usually handle these tasks independently with optimal choice of several parameters.

\underline{\it Intra-view Similarities.}
Intra-view similarity should reflect spatial arrangement of feature descriptors in each view. 
Based on the \textit{self-expressiveness property}~\cite{elhamifar2013sparse1} of an individual view, each shot can be sparsely represented by a small subset of shots that are highly correlated in the dataset. Mathematically, for $k$-th view, it can be represented as 
\begin{equation} 
\begin{gathered}
\label{eq:intra-view similarities}
x_i^{(k)}=X^{(k)}c_i^{(k)}, \ c_{ii}^{(k)}=0,
\end{gathered}
\end{equation} 
where $c_i^{(k)} = [c_{i1}^{(k)},c_{i2}^{(k)},...,c_{iN_k}^{(k)}]^T$, and the constraint $c_{ii}^{(k)}=0$ eliminates the trivial solution of representing a shot with itself.
The coefficient vector $c_i^{(k)}$ should have nonzero entries for a few shots that are correlated and zeros for the rest. 
However, in (\ref{eq:intra-view similarities}), the representation of $x_i$ in the dictionary $X$ is not unique in general. 
Since we are interested in efficiently finding a nontrivial sparse representation of $x_i$, we consider the tightest convex relaxation of the $\ell_0$ norm, i.e.,
\begin{equation}
\begin{gathered}
\label{eq:intra-view similarities1}
\text{min} \ \   ||c_i^{(k)}||_1 \ \  \text{s.t.}  \ \ x_i^{(k)}=X^{(k)}c_i^{(k)}, \ c_{ii}^{(k)}=0,
\end{gathered}
\end{equation} 
It can be rewritten in matrix form for all shots in a view as    
\begin{equation}
\begin{gathered}
\label{eq:intra-view similarities2}
\text{min} \ \   ||C^{(k)}||_1 \ \  \text{s.t.}  \  X^{(k)}=X^{(k)}C^{(k)}, \ \text{diag}(C^{(k)}) = 0,
\end{gathered}
\end{equation}
where $C^{(k)}= [c_1^{(k)},c_2^{(k)},...,c_{N_k}^{(k)}]$ is the sparse coefficient matrix whose $i$-th column corresponds to the sparse representation of the shot $x_i^{(k)}$. 
The coefficient matrix obtained from the above $\ell_1$ sparse optimization essentially characterizes the shot correlations and thus it is natural to utilize as intra-view similarities. 
This provides an immediate choice of the intra-view similarity matrix as $C_{intra}^{(k)} = |C^{(k)}|^T$ where $i$-th row of matrix $C_{intra}^{(k)}$ represents the similarities between the $i$-th shot to all other shots in the view.

\underline{\it Inter-view Similarities.}
Since all cameras are focusing on roughly the same fovs from different viewpoints, all views have apparently a single underlying structure. 
Following this assumption in a multi-view setting, we find the correlated shots across two views on solving a similar $\ell_1$ sparse optimization like in intra-view similarities. 
Specifically, we calculate the pairwise similarity between $m$-th and $n$-th view by solving the following optimization problem:
\begin{equation}
\begin{gathered}
\label{eq:inter-view similarities}
\text{min} \ \   ||C^{(m,n)}||_1 \ \  \text{s.t.}  \ \ X^{(m)}=X^{(n)}C^{(m,n)},
\end{gathered}
\end{equation}
where $C^{(m,n)} \in \mathbb{R}^{N_n \times N_m}$is the sparse coefficient matrix whose $i$-th column corresponds to the sparse representation of the shot $x_i^{(m)}$ using the dictionary $X$.
Ideally, after solving the proposed optimization problem in (\ref{eq:inter-view similarities}), we obtain a sparse representation for a shot in $m$-th view whose nonzero elements correspond to shots from $n$-th view that belong to the same subspace. 
Finally, the inter-view similarity matrix between $m$-th and $n$-th view can be represented as $C_{inter}^{(m,n)} = |C^{(m,n)}|^T$ where $i$-th row of matrix $C_{inter}^{(m,n)}$ represent similarities between $i$-th shot of $m$-th view and all other shots in the $n$-th view. 

\underline{\it Objective Function.}
The aim of embedding is to correctly match the proximity score between two shots $x_i$ and $x_j$ to the score between corresponding embedded points $y_i$ and $y_j$ respectively.
Motivated by this observation, we reach the following objective function on the embedded points $Y$.
\begin{equation*}
\begin{gathered}
\label{eq:Completeequation}
\mathcal{J}(Y^{(1)},...,Y^{(K)}) = \sum_{k}\mathcal{J_\text{intra}}(Y^{(k)}) + \sum_{m,n \atop m\neq n}\mathcal{J_\text{inter}}(Y^{(m)},Y^{(n)})
\end{gathered}
\end{equation*}
\begin{equation}
\begin{gathered}
\label{eq:Completeequation1}
= \sum_k\sum_{i,j} ||y_{i}^{(k)}-y_{j}^{(k)}||^{2}{C_{intra}^{(k)}(i,j)} + \\ \sum_{m,n \atop m\neq n}\sum_{i,j} ||y_{i}^{(m)}-y_{j}^{(n)}||^{2}{C_{inter}^{(m,n)}(i,j)}
\end{gathered}
\vspace{-1mm}
\end{equation}
where $k$, $m$ and $n = 1,\cdots,K$. $\mathcal{J_\text{intra}}(Y^{(k)})$ is the cost of preserving local correlations within $X^{(k)}$ and $\mathcal{J_\text{inter}}(Y^{(m)},Y^{(n)})$ is the cost of preserving correlations between $X^{(m)}$ and $X^{(n)}$.
The first term says that if two shots $(x_i^{(k)},x_j^{(k)})$ of a view are similar, which happens when ${C_{intra}^{(k)}(i,j)}$ is larger, their locations in the embedded space, $y_i^{(k)}$ and $y_j^{(k)}$ should be close to each other. Similarly, the second term tries to preserve the inter-view correlations by bringing embedded points $y_i^{(m)}$ and $y_i^{(n)}$ close to each other if the pairwise proximity score ${C_{inter}^{(m,n)}(i,j)}$ is high. 
Problem (\ref{eq:Completeequation1}) can be rewritten using one similarity matrix defined over the whole set of video shots as
\begin{equation}
\begin{gathered}
\label{eq:Completeequation2}
\mathcal{J}(Y) = \sum_{m,n}\sum_{i,j}||y_{i}^{(m)}-y_{j}^{(m)}||^{2}{C_{total}^{(m,n)}(i,j)} 
\end{gathered} \vspace{-1mm}
\end{equation}
where the total similarity matrix is defined as 
\begin{equation}
\label{eq:Total matrix}
C_{total}^{(m,n)}(i,j) =
\begin{cases}
C_{intra}^{(k)}(i,j) & \text{if } \, m = n = k \\
C_{inter}^{(m,n)}(i,j) & \text{otherwise}
\end{cases}
\vspace{-1mm}
\end{equation}

This construction defines a $N\times N$ similarity matrix where the diagonal blocks represent the intra-view similarities and off-diagonal blocks represent inter-view similarities.
Note that an interesting fact about our total similarity matrix construction in (\ref{eq:Total matrix}) is that since each $\ell_1$ optimization is solved individually, a fast parallel computing strategy can be easily adopted for efficiency. 
However, the matrix in (\ref{eq:Total matrix}) is not symmetric since in $\ell_1$ optimization (\ref{eq:intra-view similarities1}, \ref{eq:inter-view similarities}), a shot $x_i$ can be represented as a linear combination of some shots including $x_j$, but $x_i$ may not be present in the sparse representation of $x_j$. 
But, ideally, a similarity matrix should be symmetric in which shots belonging to the same subspace should be connected to each other. 
Hence, we reformulate (\ref{eq:Completeequation2}) with a symmetric similarity matrix $W=C_{total}+C_{total}^T$ as
\begin{equation}
\begin{gathered}
\label{eq:Completeequation3}
\mathcal{F}(Y) = \sum_{m,n}\sum_{i,j}||y_{i}^{(m)}-y_{j}^{(m)}||^{2}{W^{(m,n)}(i,j) 
}
\end{gathered} \vspace{-1mm}
\end{equation}   

With the above formulation, we make sure that two shots $x_i$ and $x_j$ get connected to each other either $x_i$ and $x_j$ is in the sparse representation of the other. Furthermore, we normalize $W$ as $w_i \leftarrow w_i/||w_i||_{\infty}$ to make sure the weights in the similarity matrix are of same scale.

Given this construction, problem (\ref{eq:Completeequation3}) reduces to the Laplacian embedding~\cite{belkin2001laplacian} of shots defined by the similarity matrix $W$. 
So, the optimization problem can be written as
\begin{equation}
\label{eq:prox_op}
Y^* = \underset{Y,YY^{T}=I}{\operatorname{argmin}}\,\, tr\big(YLY^{T} \big) 
\vspace{-1mm}
\end{equation} 
where $L$ is the graph Laplacian matrix of $W$ and $I$ is an identity matrix.
Minimizing (\ref{eq:prox_op}) is a generalized eigenvector problem and the optimal solution can be obtained by the bottom $d$ nonzero eigenvectors. 
Note that our approach is agnostic to the choice of embedding algorithms.
Our method is based on graph Laplacian because it is one of the state-of-the-art methods in characterizing the manifold structure and performs satisfactorily well in several vision and multimedia applications~\cite{gao2010local,lu2013image,nie2011unsupervised}. 

\vspace{-3mm} 
\subsection{Sparse Representative Selection}
\label{sec:Summarization}
Once the embedding is obtained, our next goal is to find an optimal subset of all the embedded shots, such that each shot can be described as weighted linear combination of a few of the shots from the subset. 
The subset is then referred as the informative summary of the multi-view videos. 
In particular, we are trying to represent the multi-view videos by selecting only a few representative shots.
Therefore, our natural goal is to establish a shot level sparsity which can be induced by performing $\ell_1$ regularization on rows of the sparse coefficient matrix~\cite{Scalable2012,Ehsan2012}.
By introducing row sparsity regularizer, the summarization problem can now be succinctly formulated as
\begin{equation}
	\begin{gathered}
		\label{eq:summ equation}
		\min_{Z\in \mathbb{R}^{N \times N}} \ \   ||Z||_{2,1} \ \  \text{s.t.}  \ \ Y=YZ
	\end{gathered} \vspace{-1mm}
\end{equation}
where $\lVert {Z \rVert}_{2,1}\triangleq \sum_{i=1}^{N}\lVert {z^i \rVert}_2$ is the row sparsity regularizer i.e., sum of $l_2$ norms of the rows of $Z$.
The self-expressiveness constraint ($Y=YZ$) in summarization is logical as the representatives for summary should come from the original frame set. Using Lagrange multipliers, (\ref{eq:summ equation}) can be written as  
\begin{equation}
\small
\begin{gathered}
\label{eq:summ equation1}
\min_{Z}\ \lVert {Y - YZ \rVert}^2_{F} + \lambda \lVert {{Z}\rVert}_{2,1} 
\end{gathered}
\end{equation}  
where $\lambda$ is a regularization parameter that balances the weight of the two terms.
Once (\ref{eq:summ equation1}) is solved, the representative shots are selected as the points whose corresponding $||z^{i}||_2 \neq 0$.
 
\underline{\it Remark 1.} Notice that both sparse optimizations in (\ref{eq:intra-view similarities2}) and (\ref{eq:summ equation}) look similar; however, the nature of sparse regularizer in both formulations are completely different. 
In (\ref{eq:intra-view similarities2}), the objective of $\ell_1$ regularizer is to induce element wise sparsity in a column whereas in (\ref{eq:summ equation}), the objective of $\ell_{2,1}$ regularizer is to induce row level sparsity in a matrix. 

\underline{\textit{Remark 2.}} Given non-uniform length of shots, (\ref{eq:summ equation1}) can be modified to a weighted $\ell_{2,1}$-norm based objective to consider length of video shots while selecting representatives as 
\begin{equation}
\small
\begin{gathered}
\label{eq:summ equation3}
\min_{Z}\ \lVert {Y - YZ \rVert}^2_{F} + \lambda \lVert {{QZ}\rVert}_{2,1} 
\end{gathered} \vspace{-1mm}
\end{equation} 
where $Q=[diag(q)]$ and $q \in \mathbb{R}^N$ represent the temporal length of each video shot. It is easy to see that problem (\ref{eq:summ equation3}) favors selection of shorter video shots by assigning a lower score via $Q$. In other words, problem (\ref{eq:summ equation3}) tries to minimize the number of shots by considering the temporal length of video shots, such that the overall objective turns to minimizing the length of the final summary. 

\vspace{-3mm}
\subsection{Joint Embedding and Sparse Representative Selection}
\label{sec:Joint}

We now discuss our proposed method to jointly optimize the multi-view video embedding and sparse representation to select a diverse set of representative shots. Specifically, the performance of sparse representative selection is largely determined by the effectiveness of graph Laplacian in embedding learning. Hence, it is a natural choice to adaptively change the graph Laplacian with respect to the following sparse representative selection, such that the embedding can not only characterizes the manifold structure, but also indicates the requirements of sparse representative selection.

By combining the objective functions (\ref{eq:prox_op}) and (\ref{eq:summ equation1}), the joint objective function becomes:
\begin{equation}
\small
\begin{gathered}
\label{eq:joint} 
\min_{{Y}, {Z}, {Y}{Y}^T={I}}  tr({Y}{L}{Y}^T) +\alpha \big( ||{Y} - {Y}{Z}||_{F}^2 + \lambda ||Z||_{2,1}\big) 
\end{gathered} \vspace{-1mm}
\end{equation}
where $\alpha > 0$ is a trade-off parameter between the two objectives. The first term of the cost function projects the input data into a latent embedding by capturing the meaningful structure of data, whereas the second term helps in selecting a robust set of representatives by minimizing the reconstruction error and the sparsity. Note that the proposed method is also computationally efficient as the sparse representative selection is done in the low-dimensional space by discarding the irrelevant part of a data point represented by a high-dimensional feature, which can derail the representative selection process.  

\vspace{-2mm}
\section{Optimization}
\label{sec:opt} 
The optimization problem in (\ref{eq:joint}) is non-smooth and non-convex. Solving it is thus more difficult due to the non-smooth $\ell_{2,1}$ norm and the additional embedding variable $Y$. 
Half-quadratic optimization techniques~\cite{he20122,he2014half} have shown to be effective in solving these sparse optimizations in several vision and multimedia applications~\cite{wang2013learning,wang2015robust,peng2016robust,lu2013correntropy}. 
Motivated by such methods, we devise an iterative algorithm to efficiently solve (\ref{eq:joint}) by minimizing its augmented function alternatively\footnote{We solve all the sparse optimization problems using Half-quadratic optimization techniques~\cite{he20122,he2014half}. Due to space limitation, we only present the optimization procedure to solve (\ref{eq:joint}). However, the same procedure can be easily extended to solve other sparse optimizations (\ref{eq:intra-view similarities2},~\ref{eq:inter-view similarities}).}.
Specifically, if we define $\phi(x)=\sqrt{x^2+\epsilon}$ with $\epsilon$ being a constant, we can transform $\lVert {{Z}\rVert}_{2,1}$ to $\sum_{i=1}^{n}\sqrt{||{z}^{i}||_2^2+\epsilon}$, according to the analysis of $\ell_{2,1}$-norm in~\cite{he20122,lu2013correntropy}. With this transformation, we can optimize (\ref{eq:joint}) efficiently in an alternative way as follows.

According to the half-quadratic theory~\cite{he20122,he2014half,geman1992constrained}, the augmented cost-function of (\ref{eq:joint}) can be written as 
\begin{equation}
\begin{gathered}
\label{eq:aug} 
\min_{{Y}, {Z}, {Y}{Y}^T={I}} tr({Y}{L}{Y}^T) +\alpha \big(||{Y} -{Y}{Z}||_{F}^2 + \lambda tr({Z}^T{P}{Z})\big) 
\end{gathered} \vspace{-1mm}
\end{equation}
where $P\in \mathbb{R}^{N\times N}$ is a diagonal matrix, and the corresponding $i$-th element is defined as
\begin{equation}
\begin{gathered}
\label{eq:diagmat2}
{P}_{i,i}=\dfrac{1}{2\sqrt{||{z}^{i}||^2_2+\epsilon}}
\end{gathered}
\vspace{-1mm}
\end{equation}
where $\epsilon$ is a smoothing term, which is usually set to be a small constant value. With this transformation, note that the problem (\ref{eq:aug}) is convex separately with respect to ${Y}, {Z}$, and ${P}$. Hence, we can solve (\ref{eq:aug}) alternatively with the following three steps with respect to ${Z}, {Y}$, and ${P}$, respectively.
\newline 
(1) \underline{\it Solving for ${Z}$}: For a given ${P}$ and ${Y}$, solve the following objective to estimate ${Z}$:    
\begin{equation}
\begin{gathered}
\label{eq:multiRSRS1refZ}
\min_{{Z}}  \alpha \big( tr(({Y}-{Y}{Z})({Y}-{Y}{Z})^T) + \lambda tr({Z}^T{P}{Z})\big)
\end{gathered} \vspace{-1mm}
\end{equation}

By setting derivative of (\ref{eq:multiRSRS1refZ}) with respect to ${Z}$ to zero, the optimal solution can be computed by solving the following linear system.
\begin{equation}
\begin{gathered}
\label{eq:multiRSRS1refZ1}
({Y}^T{Y}+\lambda{P}){Z} = {Y}^T{Y}
\end{gathered} \vspace{-2mm}
\end{equation} 
\newline 
(2) \underline{\it Solving for ${Y}$}: For a given ${P}$, and ${Z}$, solve the following objective to estimate ${Y}$:
\begin{equation}
\begin{gathered}
\label{eq:multiRSRS1refY}
\min_{{Y},{Y}{Y}^T={I}} tr({Y}{L}{Y}^T)+ \alpha  tr(({Y}-{Y}{Z})({Y}-{Y}{Z})^T)\\
=\min_{{Y},{Y}{Y}^T={I}} tr({Y}({L}+\alpha({I}-2{Z}+{Z}{Z}^T)){Y}^T)
\end{gathered} \vspace{-1mm}
\end{equation}
Eq.~\ref{eq:multiRSRS1refY} can be solved by eigen-decomposition of the matrix $({L}+\alpha({I}-2{Z}+{Z}{Z}^T))$. We pick up the eigenvectors corresponding to the $d$ smallest eigenvalues.
\newline
(3) \underline{\it Solving for ${P}$}: When ${Z}$ is fixed, we can update ${P}$ by employing the formulation in Eq.~\ref{eq:diagmat2} directly.

We continue to alternately solve for ${Z}, {Y}$, and ${P}$ until a maximum number of iterations is reached or a predefined threshold is reached. Since the alternating minimization can stuck in a local minimum, it is important to have a sensible initialization.
We initialize $Y$ by solving (\ref{eq:prox_op}) using an Eigen decomposition and ${P}$ by an identity matrix. Experiments show that the alternating minimization converges fast by using this kind of initialization. In practice, we monitor the convergence within less than 25 iterations.
Therefore, the proposed method can be applied to large scale problems in practice.

\vspace{-2mm}
\section{Summary Generation} 
\label{sec:sumg}

Above, we described how we compute the optimal sparse coefficient matrix $Z$ by jointly optimizing the multi-view embedding learning and sparse representative selection. We follow the following rules to extract a multi-view summary:

\vspace{1mm}
(i) We first generate a weight curve using $\ell_2$ norms of the rows in $Z$ since it provides information about the relative importance of the representatives for describing the whole videos. 
More specifically, a video shot with higher importance takes part in the reconstruction of many other video shots, hence its corresponding row in $Z$ has many nonzero elements with large values.
On the other hand, a shot with lower importance takes part in reconstruction of fewer shots in the whole videos, hence, its corresponding row in $Z$ has a few nonzero elements with smaller values. Thus, we can generate a weight curve, where the weight measures the confidence of the video shot to be included in the final video summary.

\vspace{1mm}
(ii) We detect local maxima from the weight curve,
then extract an optimal summary of specified length from the local maximums constrained by the weight value and full sequence coverage assumption.
Note that the shots with low or zero weights cannot be inserted into final video summary. 
Furthermore, the weigh curve in our framework allows users to choose different number of shots in summary without incurring additional computational cost. In contrast, many other multi-view video summarization methods need to preset the number of video shots that should be included in the final summary and any change will result in a re-calculation. Therefore, the proposed approach is scalable in generating summaries of different lengths and hence provides more flexibility for practical applications. More details on the summary length and scalability are included in experiments.

%% file: Experiments.tex
\begin{table} [t]
	\centering
	\tiny
	\caption{Dataset Statistics}
	\vspace{-2mm}
	\label{tab:Dataset}
	\begin{tabulary}{1.1\linewidth}{|p{12mm}||p{8mm}|P{24.5mm}|P{8mm}|P{12mm}|}
		\hline
		\textbf{\texttt{Datasets}}& \textbf{\texttt{\# Views}} & \textbf{\texttt{Total Durations (Mins.)}} & \textbf{\texttt{Settings}} & \textbf{\texttt{Camera Type}} \\
		\hhline{|=|=|=|=|=|}
		\textbf{\texttt{Office}} &	\centering 4	& 46:19	& Indoor	& Fixed	\\
		\textbf{\texttt{Campus}} & \centering4 & 56:43	& Outdoor	& Non-fixed	\\
		\textbf{\texttt{Lobby}} &\centering3	& 24:42	& Indoor	& Fixed	\\
		
		\textbf{\texttt{Road}} &\centering3	& 22:46	& Outdoor	& Non-fixed	\\
		\textbf{\texttt{Badminton}} &\centering3	& 15:07	& Indoor	& Fixed		\\
		\textbf{\texttt{BL-7F}} &\centering19	& 136:10	& Indoor	& Fixed	\\
		\hline 		
	\end{tabulary} 
	\vspace{-3mm}
\end{table} 
In this section, we present various experiments and comparisons
to validate the effectiveness and efficiency of our proposed algorithm in summarizing multi-view videos.
\vspace{-3mm}
\subsection{Datasets.} We conduct rigorous experiments using 6 multi-view datasets with 36 videos in total, which are from~\cite{MultiviewTMM2010,OnlineMultiview2015} (See Tab.~\ref{tab:Dataset}). The datasets are captured in both indoor and outdoor environments with overall 360 degree coverage of the scene, making it more difficult to be summarized. All these datasets are standard in multi-view video summarization and have been used by the prior works~\cite{MultiviewTMM2010,SanjaySir2015,MultiviewICIP2011}. It is important to note that experiments in our prior work~\cite{panda2016video} was limited to only 3 datasets, whereas in the current work, we conduct experiments on 6 datasets including BL-7F which is one of the largest publicly available dataset for multi-view video summarization.  

\vspace{-3mm}
\subsection{Performance Measures.} To provide an objective comparison, we compare all the approaches using three quantitative measures, including Precision, Recall and F-measure ($\frac{2\times Precision\times Recall}{Precision+Recall}$)~\cite{MultiviewTMM2010,SanjaySir2015}.
For all these metrics, the higher value indicates better summarization quality. 
We set the same summary length as in~\cite{MultiviewTMM2010} to generate our summaries and employ the ground truth of important events reported in~\cite{MultiviewTMM2010} to compute the performance measures. More specifically, the ground truth annotations contain a list of events with corresponding start and end frame for each dataset. We took an event as correctly detected if our framework produces a video shot between the start and end of the event. We follow the prior works~\cite{MultiviewTMM2010,SanjaySir2015,OnlineMultiview2015} and consider an event to be redundant if we detect the event simultaneously from more than one camera. Such an evaluation setting gives a fair comparison with the previous state-of-the-art methods~\cite{MultiviewTMM2010,MultiviewICIP2011,SanjaySir2015,panda2016embedded,panda2016video}.

\begin{table*} [t]
	\scriptsize
	\centering
	\caption{Performance comparison with several baselines including both single and multi-view methods applied on the three multi-view datasets. \textbf{P}: Precision in percentage, \textbf{R}: Recall in percentage and $\bf{F}$: {F}-measure. Ours perform the best.
	} 	
	\vspace{-2mm}
	\label{tab:Multi-view Comparison Table}
	\begin{tabulary}{1.1\linewidth}{|p{30mm}|P{7mm}|P{7mm}|P{8mm}|P{7mm}|P{7mm}|P{8mm}|P{7mm}|P{7mm}|P{8mm}|P{20mm}|}
		\hline
		
		& \multicolumn{3}{c|}{\textbf{Office}} &\multicolumn{3}{c|}{\textbf{Campus}}&\multicolumn{3}{c|}{\textbf{Lobby}} & \\
		\cline{2-10}
		
		\textbf{Methods} &\textbf{P}& \textbf{R} & $\bf{F}$ &\textbf{P}& \textbf{R} & $\bf{F}$ &\textbf{P}& \textbf{R} & $\bf{F}$ & \textbf{Reference}\\
		\hhline{|=|=|=|=|=|=|=|=|=|=|=|}
		\texttt{Attention-Concate}	& 100	& 46	& 63.01 & 40	& 28	& 32.66 & 100	& 70  & 82.21 & TMM2005~\cite{Att2005}	\\
		\texttt{Sparse-Concate}	& 100 	& 50	&66.67  & 56	& 55	&55.70  & 91	& 70 & 78.95& TMM2012~\cite{Scalable2012}\\
		\texttt{Concate-Attention}	& 100	& 38	& 55.07 & 56	& 48	& 51.86 & 95	& 72 &  81.98 & TMM2005~\cite{Att2005}\\
		\texttt{Concate-Sparse}	& 93	& 58 & 71.30 & 56	& 62 & 58.63 &86 	& 70 & 77.18	&TMM2012~\cite{Scalable2012}\\ 
		\texttt{Graph}	& 100	& 26	& 41.26 & 50	& 48	& 49.13  & 100	& 58 & 73.41& TCSVT2006~\cite{peng2006clip}\\
		\texttt{RandomWalk}	& 100	& 61	& 75.77 & 70	& 55	& 61.56 & 100	& 77 & 86.81	& TMM2010~\cite{MultiviewTMM2010} \\ 
		\texttt{RoughSets} 	& 100	& 61	& 75.77 & 69	& 57	& 62.14 & 97	& 74 & 84.17 & ICIP2011~\cite{MultiviewICIP2011} \\
		\texttt{BipartiteOPF} & 100	& 69	&  81.79 & 75	& 69	& 71.82 & 100	& 79 & 88.26	&TMM2015~\cite{SanjaySir2015}\\ 
		\texttt{Ours}	& \textbf{100}	& \textbf{81} & \textbf{89.36} & \textbf{84}	& \textbf{72} & \textbf{77.78} & \textbf{100}	& \textbf{86} & \textbf{92.52}& Proposed\\ \hline 
		
	\end{tabulary} \vspace{-2mm}
\end{table*}

\vspace{-2mm}
\subsection{Experimental Settings.}
We maintain the following conventions during all our experiments.
(i) All our experiments are based on unoptimized MATLAB codes on a desktop PC with Intel(R) core(TM) i7-4790 processor with 16
GB of DDR3 memory. We used a NVIDIA Tesla K40 GPUs to extract the C3D features. (ii) Each feature descriptor is $L_2$-nominalized. (iii) Determining the intrinsic dimensionality of the embedding is an open problem in the field of manifold learning. One common way is to determine it by grid search. We determine it as in most traditional approaches, such as~\cite{cai2007spectral}. (iv) The sparsity regularization parameter $\lambda$ is computed as $\lambda_0 / \rho$ and $\lambda_0$ is analytically computed from the embedded points~\cite{Ehsan2012}, (v) We empirically set $\alpha$ to 0.05 and kept fixed for all results. 
\vspace{-3mm}
\subsection{\it Comparison with State-of-the-art Multi-view Methods.}
\underline{\it Goal.} This experiment aims at evaluating our approach compared to the state-of-the-art multi-view
summarization methods presented in the literature.

\underline{\it Compared Methods.} We contrast our approach with several state-of-the-art methods which are specifically designed for multi-view video summarization as follows.

$\bullet$ \texttt{RandomWalk}~\cite{MultiviewTMM2010}. The method first create a spatio-temporal shot graph and then use random walk as a clustering algorithm over the graph to extract multi-view summaries.

$\bullet$ \texttt{RoughSets}~\cite{MultiviewICIP2011}. The method first adopt a SVM classifier as the key frame abstraction process and then applies rough set to remove similar frames.

$\bullet$ \texttt{BipartiteOPF}~\cite{SanjaySir2015}. This method first uses a bipartite graph matching to model the inter-view correlations and then applies optimum path forest clustering on the refined adjacency matrix to generate multi-view summary.

$\bullet$ \texttt{GMM}~\cite{OnlineMultiview2015}. An online Gaussian mixture model clustering is first applied on each view independently and then a distributed view selection algorithm is adopted to remove the content redundancy in the inter-view stage.

\underline{\it Implementation Details.} To report existing methods results, we use prior published numbers when possible. In particular, for the multi-view summarization methods (\texttt{RandomWalk}, \texttt{BipartiteOPF} and \texttt{GMM}), we report the available results from the corresponding papers and implement \texttt{RoughSets} ourselves using the same video representation as the proposed one and tune their parameters to have the best performance.  

\begin{figure} 
	\centering
	{
		\includegraphics[width=1\linewidth]{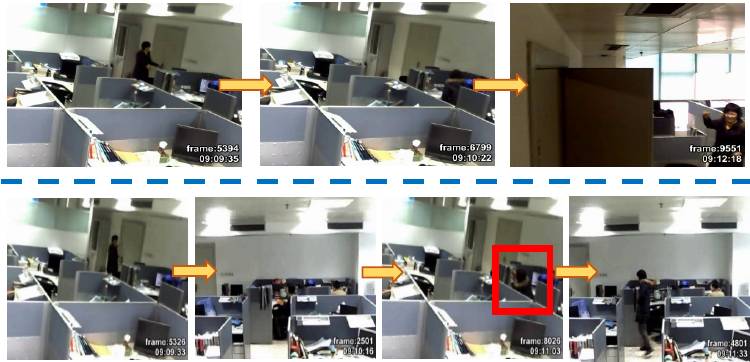}}
	\vspace{-6mm}
	\caption
	{Sequence of events detected related to activities of a member $(A_0)$ inside the Office dataset. Top row: Summary produced by method~\cite{MultiviewTMM2010}, and Bottom row: Summary produced by our approach. 
		Sequence of events detected in top row: 1st: $A_0$ enters the room, 2nd: $A_0$ sits in cubicle 1, 3rd: $A_0$ leaves the room. Sequence of events detected in bottom row: 1st: $A_0$ enters the room, 2nd: $A_0$ sits in cubicle 1, 3rd: $A_0$ is looking for a thick book to read (as per the ground truth in~\cite{MultiviewTMM2010}), and 4th: $A_0$ leaves the room. 
		The event of looking for a thick book to read (as per the ground truth in~\cite{MultiviewTMM2010}) is missing in the summary produced by method~\cite{MultiviewTMM2010} where as it is correctly detected by our approach (3rd frame: bottom row). This indicates our method captures video semantics in a more informative way compared to~\cite{MultiviewTMM2010}.   
	}
	\label{fig:Event Detection} 
	\vspace{-3mm}
\end{figure}  

\begin{figure*}[h]
	\centering
	\begin{tabular}{c}
		\includegraphics[width=0.98\linewidth]{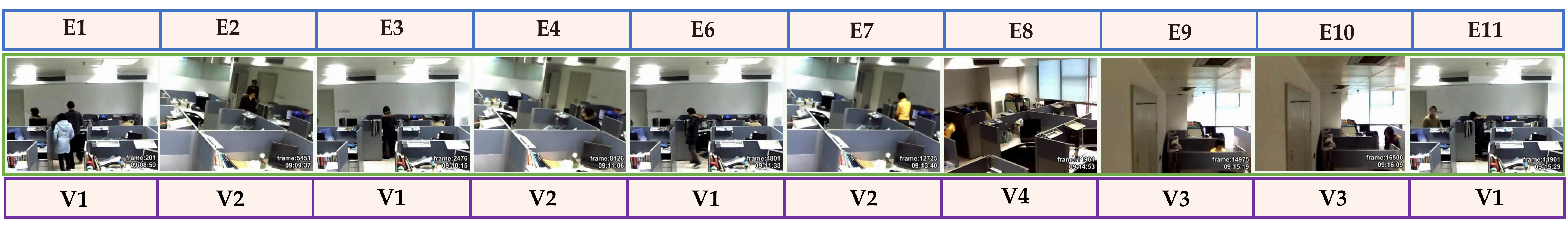}
	\end{tabular}
	\vspace{-4mm}
	\caption
	{Summarized events for the Office dataset. Each event is represented by a key frame and is associated with two numbers, one above and below of the key frame. Numbers above the frame (\texttt{E1}, $\cdots$, \texttt{E26}) represent the event number whereas the numbers below (\texttt{V1}, $\cdots$, \texttt{V4}) indicate the view from which the event is detected.    
	Limited to the space, we only present 10 events arranged in temporal order, as per the ground truth in~\cite{MultiviewTMM2010}. 
	}
	\label{fig:OfficeEvents}
	\vspace{-2mm}
\end{figure*}

\underline{\it Results.} Table~\ref{tab:Multi-view Comparison Table} shows the results on three multi-view datasets, namely Office, Campus and Lobby datasets. We have the following key observations from Table~\ref{tab:Multi-view Comparison Table}: (i) Our approach produces summaries with same precision as \texttt{RandomWalk} and \texttt{BipartiteOPF} for both Office and Lobby datasets. However, the improvement in recall value indicates the ability of our method in keeping more important information in the summary compared to both of the approaches.
As an illustration, in Office dataset, the event of looking for a thick book by a member while present in the cubicle is absent in the summary
produced by \texttt{RandomWalk} whereas it is correctly detected by our proposed method. Fig.~\ref{fig:Event Detection} in this connection explains
the whole sequence of events detected using our approach
as compared to \texttt{RandomWalk}.
(ii) For all methods, including \texttt{Ours}, performance on Campus dataset is not that good as compared to the other datasets.
This is obvious since the Campus dataset contains many
trivial events as it was captured in an outdoor environment, thus making the summarization more difficult. Nevertheless,
for this challenging dataset, F-measure of our approach
is about 6\% better than that of the recent \texttt{BipartiteOPF}. (iii) Table~\ref{tab:Multi-view Comparison Table} also reveals that for all three datasets, recall is generally low compared to precision because users usually prefer to select more extensive summaries in ground truth, which can be verified from the ground truth events from~\cite{MultiviewTMM2010}. As a result, number of events in ground truth increases irrespective of their information content.  (iv) Overall, on the three datasets, our approach outperforms all compared methods in terms of F-measure. This corroborates the fact that the proposed approach produces informative
multi-view summaries in contrast to the state-of-the-art methods (See Fig.~\ref{fig:OfficeEvents} for an illustrative example). 

Table~\ref{tab:BL-7F} shows results of our method on a larger and more complex BL-7F dataset captured with 19 surveillance cameras in the 7th floor of the BarryLam Building in National Taiwan University~\cite{OnlineMultiview2015}. From Table~\ref{tab:BL-7F}, it is clearly evident that our approach significantly outperforms the recent method \texttt{GMM} in generating more informative multi-view summaries. The F-measure of our method is about 11\% better than that of \texttt{GMM}~\cite{OnlineMultiview2015}. This indicates that the proposed method
is very effective and can be applied to large scale problems in practice. We follow the evaluation strategy of~\cite{OnlineMultiview2015} and compute the performance measures in the unit of frames instead of events as in Table~\ref{tab:Multi-view Comparison Table} to make a fair comparison with the \texttt{GMM} baseline.   

\begin{table} [t]
	\centering
	\tiny
	\caption{Performance Comparison with \texttt{GMM} baseline on BL-7F Dataset}
	\vspace{-2mm}
	\label{tab:BL-7F}
	\begin{tabulary}{1.1\linewidth}{|p{12mm}||p{10mm}|P{10mm}|P{15mm}|P{12mm}|}
		\hline
		\textbf{\texttt{Methods}}& \textbf{Precision(\%)} & \textbf{Recall(\%)} & \textbf{F-measure(\%)} & \textbf{Reference} \\
		\hhline{|=|=|=|=|=|}
		\texttt{GMM} &	\centering 58	& 61	& 60.00	& JSTSP2015~\cite{OnlineMultiview2015}	\\
		\texttt{Ours} & \centering 73  & 70  & 71.29 & Proposed \\
		\hline 		
	\end{tabulary} 
	\vspace{-2mm}
\end{table}

\vspace{-4mm}
\subsection{\it Comparison with Single-view Methods.}
\underline{\it Goal.} The objective of this experiment is to compare our method with some single-view summarization approaches to show their performance on multi-view videos. 
Specifically, the purpose of comparing
with single-view summarization methods is to show that techniques that attempt to find summary from
single-view videos usually do not produce an optimal set of representatives while summarizing multiple videos.  

\underline{\it Compared Methods.} We compare our approach with several baseline methods (\texttt{Attention-Concate}~\cite{Att2005}, \texttt{Sparse-Concate}~\cite{Scalable2012}, \texttt{Concate-Attention}~\cite{Att2005}, \texttt{Concate-Sparse}~\cite{Scalable2012},  \texttt{Graph}~\cite{peng2006clip}) that use single-video summarization approach over multi-view datasets to generate summary. 
Note that in the first two baselines (\texttt{Attention-Concate}, \texttt{Sparse-Concate}), a single-video summarization approach is first applied to each view and then resulting summaries are combined to form a single summary, whereas the other three baselines (\texttt{Concate-Attention}, \texttt{Concate-Sparse}, \texttt{Graph}) concatenate all the views into a single video and then apply a single-video approach to summarize multi-view videos. 
Both \texttt{Sparse-Concate} and \texttt{Concate-Sparse} baselines use (\ref{eq:summ equation1}) to summarize multi-view videos with out any embedding. The purpose of comparing with these two baseline methods is to explicitly show the advantage of our proposed multi-view embedding in generating informative and diverse summaries while summarizing multi-view surveillance videos.

\underline{\it Implementation Details.} We implement \texttt{Sparse-Concate} and \texttt{Concate-Sparse} ourselves with the same temporal segmentation and C3D feature representation as the proposed one whereas for rest of the single-view summarization methods, we report the available results from the published papers~\cite{MultiviewTMM2010,SanjaySir2015}.

\begin{figure*}[!t]
	\centering
	{
		\includegraphics[width=0.75\linewidth]{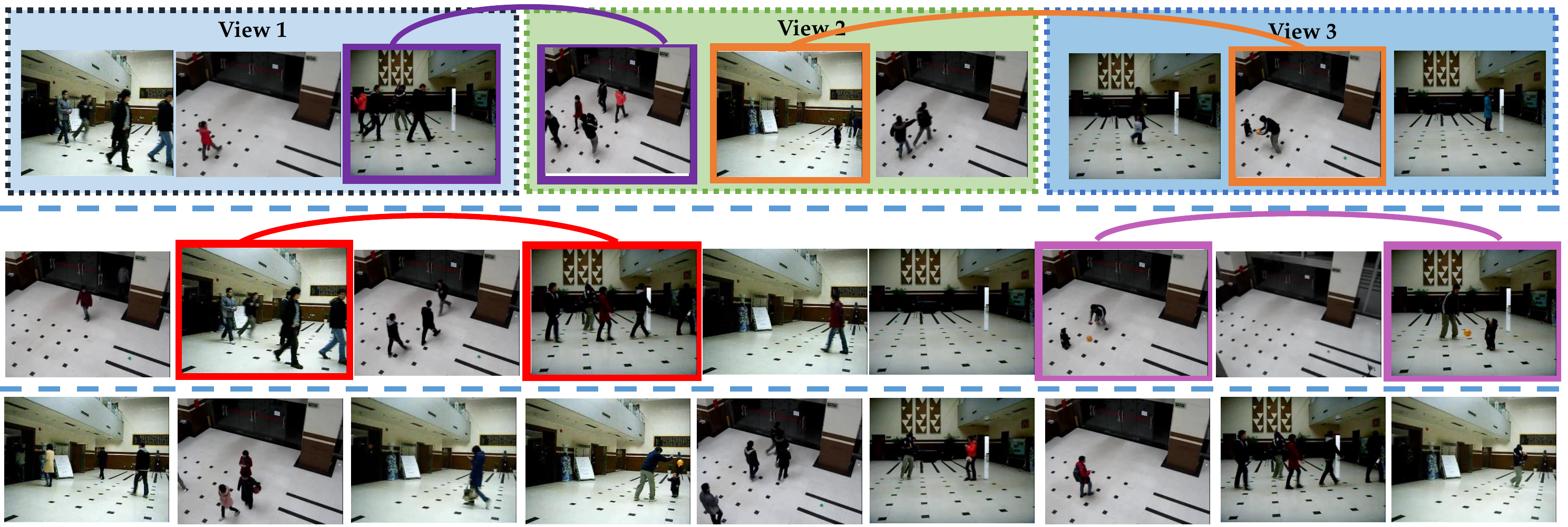}}
	\vspace{-2mm}
	\caption
	{Some summarized events for the Lobby dataset. 
		Top row: summary produced by \texttt{Sparse-Concate}~\cite{Scalable2012}, Middle row: summary produced by \texttt{Concate-Sparse}~\cite{Scalable2012}, and Bottom row: summary produced by our approach. It is clearly evident from both top and middle rows that both of the single-view baselines produce a lot of redundant events as per the ground truth~\cite{MultiviewTMM2010} while summarizing multi-view videos, however, our approach (bottom row) produces meaningful representatives by exploiting the content correlations via an embedding.
		Redundant events are marked with same color borders. 
		Note that both \texttt{Sparse-Concate} and \texttt{Concate-Sparse} summarize multiple videos without any embedding by either applying sparse representative selection to each video separately or concatenating all the videos into a single video. Best viewed in color.      
	}
	\label{fig:View-board}\vspace{-3mm}
\end{figure*}

\underline{\it Results.} We have the following key findings from Table~\ref{tab:Multi-view Comparison Table} and Fig.~\ref{fig:View-board}: (i) The proposed method significantly outperforms all the compared single-view summarization methods by a significant margin on all three datasets.
We observe that directly applying these methods to summarize multiple videos produces a lot of redundant shots which deviates from the fact that the optimal summary should be diverse and informative in describing the multi-view concepts. 
(ii) It is clearly evident from the Fig.~\ref{fig:View-board} that both of the sparse representative selection based single-view summarization methods (\texttt{Sparse-Concate} and \texttt{Concate-Sparse}) produce a lot of redundancies (simultaneous presence of most of the events) while summarizing videos on Lobby dataset. This is expected since both of the approaches fail to exploit the complicated inter-view content correlations present in multi-view videos.   
(iii) By using our multi-view video summarization method, such redundancy is largely reduced in contrast. Some events are recorded by the most informative summarized shots, while the most important events are reserved in our summaries. The proposed approach generates highly informative and diverse summary in most cases, due to its ability to jointly model multi-view correlations and sparse representative selection.    
\vspace{-3mm}
\subsection{\it Scalability in Generating Summaries.}
\begin{figure} [t]
 	\subfigure[]{
 		\label{fig:Fig3Sa}
 		\includegraphics[width=1\linewidth]{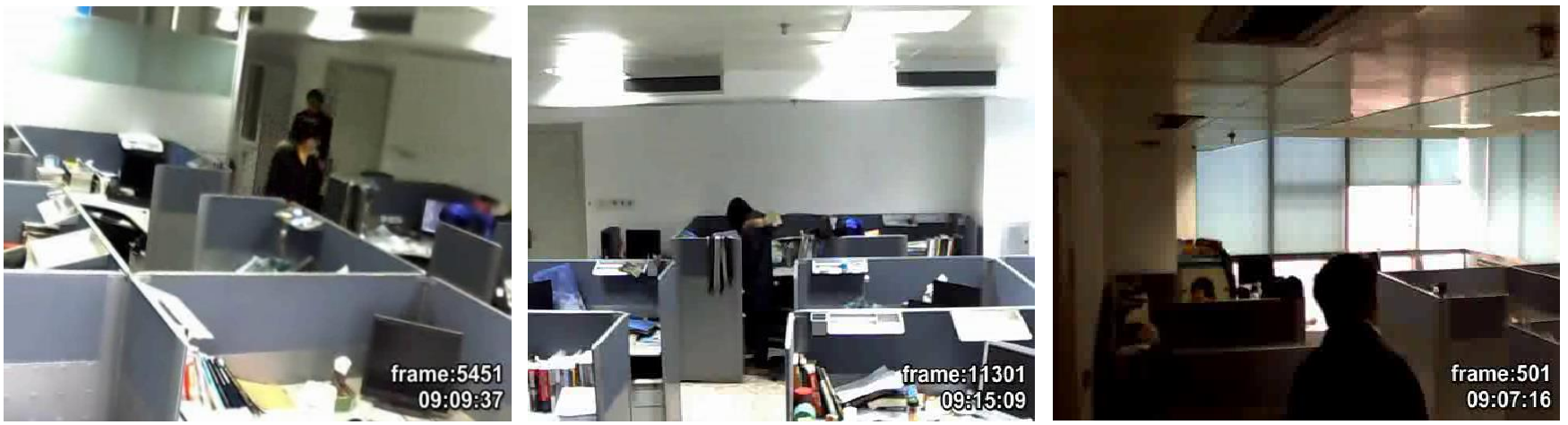}}
 	\subfigure[]{
 		\label{fig:Fig3Sb}
 		\includegraphics[width=1\linewidth]{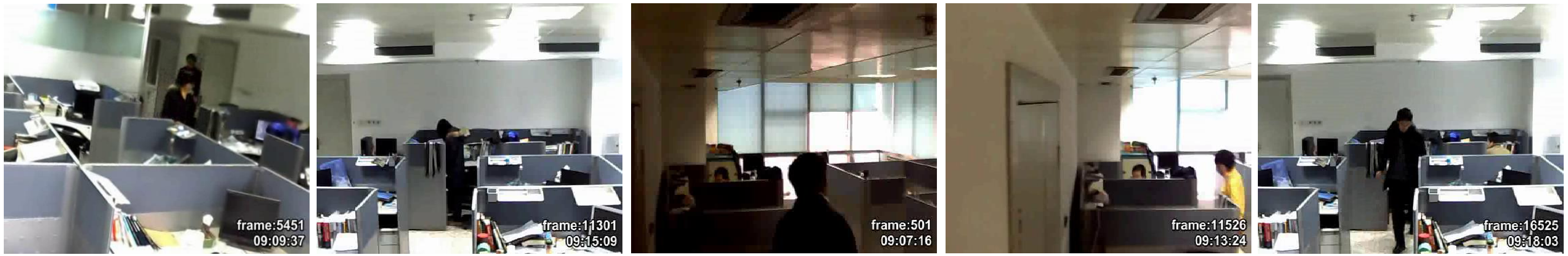}}
 	\subfigure[]{
 		\label{fig:Fig3Sc}
 		\includegraphics[width=1\linewidth]{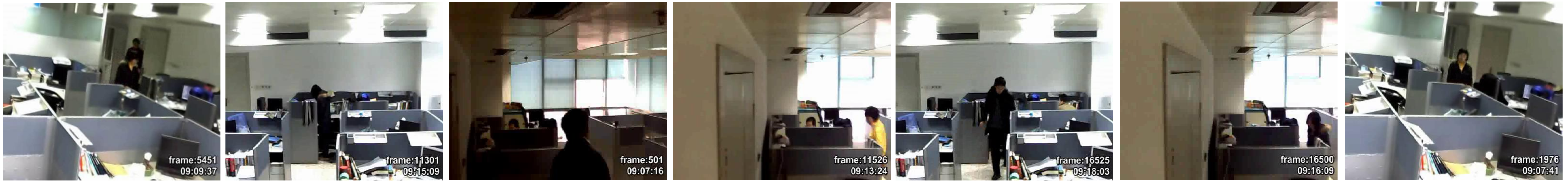}}
 	\vspace{-5mm}
 	\caption
 	{The figure shows an illustrative example of scalability in generating summaries of different length based on the user constraints for the Office dataset. Each shot is represented by a key frame and are arranged according to the $l_2$ norms of corresponding non-zero rows of the sparse coefficient matrix. ~\subref{fig:Fig3Sa}: Summary for user length request of 3, ~\subref{fig:Fig3Sb}: Summary for user length request of 5 and ~\subref{fig:Fig3Sc}: Summary for user length request of 7.
 	}
 	\label{fig:scalabilty}
 \end{figure}

Scalability in generating summaries of different length has shown to be effective while summarizing single videos~\cite{herranz2010framework,panda2014scalable}. However, most of the prior multi-view summarization methods require the number of shots to be specified before generating summaries which is highly undesirable in practical applications. Concretely speaking, the algorithm need to be rerun for each change in the number of representative shots that the user want to see in the summary. By contrast, our approach provides scalability in generating summaries of different length based on user constraints without any further
analysis of the input videos (\textit{analyze once, generate many}). This is due to the fact that non-zero rows of the sparse coefficient matrix $Z$ can generate a ranked list of representatives which can be subsequently used to provide a scalable representation in generating summaries of desired length without incurring any additional cost. Such a scalability property makes our approach more suitable in providing human-machine interface where the summary length is changed as per the user request. Fig.~\ref{fig:scalabilty} shows the generated summaries of length 3, 5 and 7 most important shots (as determined by the weight curve described in Sec.~\ref{sec:sumg}) for Office dataset.

\vspace{-3mm}
\subsection{\it Performance Analysis with Shot-level C3D Features.} We investigate the importance and reliability of the proposed video representation based on C3D features by comparing with 2D shot-level deep features, and found that the later produces inferior results, with a F-measure of 84.01\% averaged over three datasets (Office, Campus and Lobby) compared to 86.55\% by the C3D features. We utilize Pycaffe
with the VGG net pretrained model~\cite{simonyan2014very} to extract a 4096-dim feature vector of a frame and then use temporal mean pooling to compute a single shot-level feature vector, similar to C3D features described in Sec.~\ref{sec:video representation}. The spatio-temporal C3D features perform best, as they exploit the temporal aspects of activities typically shown in videos.  
 
\vspace{-3mm}
\subsection{\it Performance Analysis with Video Segmentation.} 
We examined the performance of our approach by replacing the temporal segmentation algorithm~\cite{chu2015video} by a naive approach that uniformly divides video into several segments of equal length. We use uniform segments with a length of 2 seconds and kept other components fixed while generating summaries. By using the video segmentation algorithm of~\cite{chu2015video}, the proposed approach achieves a F-measure of 86.55\% averaged over three datasets (Office, Campus and Lobby). On the other hand, with the use of uniform length segments, our approach obtains a mean F-measure 85.43\%.  
This shows that our approach is relatively robust with the change in segmentation algorithm. Note that our proposed sparse optimization is highly flexible to incorporate more sophisticated temporal segmentation algorithms, e.g.,~\cite{poleg2014temporal} in generating video summaries---we expect such advanced and complex video segmentation algorithms will only benefit our proposed approach. 
\vspace{-2mm}
\subsection{\it Performance Comparison with~\cite{panda2016video}.} 
We now compare the proposed approach with~\cite{panda2016video} to explicitly verify the effectiveness of video representation and joint optimization for summarizing multi-view videos. Table~\ref{tab:comp-ours} shows the comparison with~\cite{panda2016video} on Office, Campus and Lobby datasets. Following are the analysis of the results: (i) The proposed framework consistently outperforms ~\cite{panda2016video} on all three datasets by a margin of about 5\% in terms of F-measure (maximum improvement of 8\% in terms of precision for the office dataset). (ii) We improve around 3\% in terms of F-measure for the more challenging Campus dataset which demonstrates that the current framework is more effective in summarizing videos with outdoor scenes. (iii) We believe the best performance in the proposed framework can be
attributed to two factors working in concert: (a) more flexible and powerful video representation via C3D features, and (b) joint embedding learning and sparse representative selection. Moreover, to better understand the contribution of joint optimization, we analyzed the performance of the proposed approach with shot-level C3D features and a 2 step process similar to~\cite{panda2016video}, and found that the mean F-measure on three datasets (Office, Campus and Lobby) decreases from 86.55\% to 83.85\%. We believe this is because adaptively changing the graph Laplacian with respect to the sparse representative selection helps in better exploiting the multi-view correlations and also indicates the requirement of optimal representative shots to be included in the summary. It also important to note that the approach in~\cite{panda2016video} is limited to key frame extraction only and hence may not be suitable for many surveillance applications where video skims with motion information seems better suited for obtaining significant information in short time. 
\begin{table} [t]
	\centering
	\tiny
	\caption{F-measure Comparison with~\cite{panda2016video}}
	\vspace{-2mm}
	\label{tab:comp-ours}
	\begin{tabulary}{1.1\linewidth}{|p{12mm}||p{10mm}|P{10mm}|P{15mm}|P{12mm}|}
		\hline
		\textbf{\texttt{Methods}}& \centering\textbf{Office} & \textbf{Campus} & \textbf{Lobby} & \textbf{Reference} \\
		\hhline{|=|=|=|=|=|}
		\texttt{~\cite{panda2016video}} &	\centering 84.48	& 75.42	& 88.26	& ICPR2016~\cite{panda2016video}	\\
		\texttt{Ours} & \centering89.36 & 77.78 & 92.52& Proposed \\
		\hline 		
	\end{tabulary} 
	\vspace{-2mm}
\end{table} 
         
\vspace{-2mm}
\subsection{\it User Study.} With 5 study experts, we performed human evaluation of the generated summaries to verify the results obtained from the automatic objective evaluation with F-measure. Our objective is to understand how an user perceive the quality of the summaries according to the visual pleasantness and information content of the system generated summary. Each study expert watched the videos at 3x speed and were then shown 3 sets of summaries constructed using different methods: \texttt{RandoWalk}, \texttt{BipartiteOPF} and \texttt{Ours} for 5 datasets (Office, Campus, Lobby, Road and Badminton). Study experts were asked to rate the overall quality of each summary by assigning a rating from 1 to 10, where 1 corresponded to \enquote{The generated summary is not at all informative} and 10 corresponded to \enquote{The summary very well describes all the information present in the original videos and also visually pleasant to watch}. The summaries were shown in random order without revealing the identity of each method and the audio track was not included to ensure that the subjects chose the rating based solely on visual stimuli. The results are summarized in Table~\ref{tab:user}.
Similar to the objective evaluation, our approach significantly outperforms both of the methods (\texttt{RandomWalk}, \texttt{BipartiteOPF}). This again corroborates the fact that the proposed framework generates a more informative and diverse multi-view summary as compared to the state-of-the-art methods. Furthermore, we note that the relative rank of the different algorithms is largely preserved in the subjective user study as compared to the objective evaluation in  Table~\ref{tab:Multi-view Comparison Table}. 

\begin{table} [t]
	\centering
	\tiny
	\caption{\textbf{User Study}---Mean Expert ratings on a scale of 1 to 10. Our approach significantly outperforms other automatic methods.}
	\vspace{-2mm}
	\label{tab:user}
	\begin{tabulary}{1.1\linewidth}{|p{12mm}||p{7mm}|P{7mm}|P{7mm}|P{7mm}|P{7mm}|}
		\hline
		\textbf{\texttt{Methods}}& \centering\textbf{Office} & \textbf{Campus} & \textbf{Lobby} & \textbf{Road} & \textbf{Badminton} \\
		\hhline{|=|=|=|=|=|=|}
		\texttt{RandomWalk} &  \centering6.3	  	& 5.2	&6.6 	& 5.7 &6.5	\\
		\texttt{BipartiteOPF} & \centering7.1	 	& 5.8	& 7.4	& 6.0 &7.2	\\
		\texttt{Ours} & \centering 7.6 & 6.5 &8.2 & 6.7 &7.9 \\
		\hline 		
	\end{tabulary} 
	\vspace{-4mm}
\end{table}   

\vspace{-2mm}
\subsection{\it Discussions.} \underline{\textit{Abnormal Event Detection.}} Abnormal event detection and surveillance video summarization are two closely related problem in computer vision and multimedia. 
In a surveillance setting, where an abnormal event took place, the proposed approach can select shots to represent the abnormal event in the final summary. This is due to the fact that our approach selects representative shots from the mult-view videos such that set of videos should be reconstructed with high accuracy using the extracted summary. Specifically, the proposed approach in (\ref{eq:joint}) favors selecting a set of shots as representatives for constructing the summary which can reconstruct all the events in the input with low reconstruction error. Consider a simple example for an illustration. Let us assume a surveillance setting equipped in a place with only pedestrian traffic. People are walking as usual and suddenly, a car is speeding. In order to reconstruct the part where the car is speeding, our method will choose a few shots from this portion; otherwise the reconstruction error will be high.

\underline{\textit{Multi-View Event Capture.}} In general, the purpose of overlapping-field of view is to facilitate users to check objects/events from different angles.
For an event captured with multiple cameras having a large difference in view angles, the proposed method often selects more than one shot to represent the event in the summary. 
This is due to the fact that our approach selects representative shots from the multi-view videos such that the whole input can be reconstructed with low error.
In our experiments, we have observed a similar situation while summarizing videos on Campus dataset.
The summary produced by our approach contains three shots captured with cameras 1, 3, and 4 in an outdoor environment which essentially represent the same event (E23 in the ground truth~\cite{MultiviewTMM2010}). However, note that although including shots representing same event from more than one camera in the summary may help an user to check events from different angles, it increases the summary length which often deviates from the fact that length of the summary should be as small as possible. Thus, the objective of our current work is on generating an optimal summary that balances the two main important criteria of a good summary, i.e., maximizing the information content via representativeness and minimizing the length via sparsity.

\underline{\textit{Joint Video Segmentation and Summarization.}} Note that the proposed approach uses temporal video segmentation as a preprocessing step and then use the shot-level features to extract summaries. Our approach can be modified in two ways to optimize the temporal segmentation for the task of video summarization. 
First, involving a human in our current approach for giving feedbacks, similar to the concept of relative attributes in visual recognition~\cite{parikh2011relative} can help us in adaptively changing the shot boundaries for generating better quality summaries. Second, learning a dynamic agent using Markov decision process (MDP) for moving the shot boundaries (forward or backward with temporal increments) based on the performance of our proposed summarization algorithm is also a possibility in this regard~\cite{caicedo2015active}.
Developing an efficient framework for joint segmentation and summarization is an interesting practical problem---we leave this as future work, with no existing work, to the best of our knowledge.

%% file: Conclusions.tex
In this paper, we addressed the problem of summarizing
multi-view videos via joint embedding learning and $\ell_{2,1}$ sparse optimization.
The embedding helps is capturing content correlations in
multi-view datasets without
assuming any prior correspondence between
the individual videos. On the other hand, the sparse representative selection helps in generating multi-view summaries as per user length request without requiring additional computational cost. 
Performance comparisons on six standard multi-view datasets show marked improvement over some mono-view summarization approaches as well as state-of-the-art multi-view summarization methods. 

Moving forward, we would like to improve our method by explicitly incorporating video semantics that may require more complex model with additional techniques such as attention modeling~\cite{Att2005} or semantic feature analysis based on user preferences~\cite{lie2008video}. It is also important to note that unlike single-view videos, understanding semantics in a multi-camera environment is a challenging problem and hence, one may also require data association strategies, e.g,~\cite{chakraborty2016network} to properly exploit multi-view video semantics for the task of video summarization. Moreover in future, we would like to consider the fact that more than one camera view may be necessary to fully represent an event (e.g., due to occlusion) in a multi-view setting and hence, it may be necessary to include multiple similar shots representing same events from more than one camera for generating a good quality video summary.